\documentclass[letterpaper, 10 pt, conference]{ieeeconf}  

\IEEEoverridecommandlockouts                              

\usepackage{graphicx}
\usepackage{hyperref}
\usepackage{algpseudocode}
\usepackage{algorithm}
\usepackage{subfig}
\usepackage{float}
\usepackage{cite}
\usepackage{listings}
\usepackage{xcolor}
\usepackage{booktabs}
\usepackage{array}
\usepackage{float}

\colorlet{punct}{red!60!black}
\definecolor{background}{HTML}{EEEEEE}
\definecolor{delim}{RGB}{20,105,176}
\colorlet{numb}{magenta!60!black}
\lstdefinelanguage{json}{
    basicstyle=\normalfont\ttfamily,
    numbers=left,
    numberstyle=\scriptsize,
    stepnumber=1,
    numbersep=8pt,
    showstringspaces=false,
    breaklines=true,
    frame=lines,
    backgroundcolor=\color{background},
    literate=
     *{0}{{{\color{numb}0}}}{1}
      {1}{{{\color{numb}1}}}{1}
      {2}{{{\color{numb}2}}}{1}
      {3}{{{\color{numb}3}}}{1}
      {4}{{{\color{numb}4}}}{1}
      {5}{{{\color{numb}5}}}{1}
      {6}{{{\color{numb}6}}}{1}
      {7}{{{\color{numb}7}}}{1}
      {8}{{{\color{numb}8}}}{1}
      {9}{{{\color{numb}9}}}{1}
      {:}{{{\color{punct}{:}}}}{1}
      {,}{{{\color{punct}{,}}}}{1}
      {\{}{{{\color{delim}{\{}}}}{1}
      {\}}{{{\color{delim}{\}}}}}{1}
      {[}{{{\color{delim}{[}}}}{1}
      {]}{{{\color{delim}{]}}}}{1},
}

\title{\LARGE \bf
Task-Aware Bimanual Affordance Prediction via VLM-Guided Semantic-Geometric Reasoning

}

\author{Fabian Hahne$^1$, Vignesh Prasad$^{1, 4,5}$, Georgia Chalvatzaki$^{1, 4, 5}$, Jan Peters$^{1, 2, 3, 4, 5}$, Alap Kshirsagar$^1$
\thanks{$^1$Department of Computer Science, Technical University of Darmstadt
$^2$German Research Center for AI (DFKI) $^3$Centre for Cognitive Science, Technical University of Darmstadt $^4$Hessian Center for Artificial
Intelligence (Hessian.AI), Darmstadt $^5$Robotics Institute Germany (RIG)}
\thanks{This work was supported by the German Research Foundation (DFG) Emmy Noether Programme under Grant CH 2676/1-1 and under Germany's Excellence Strategy (EXC 3066/1 “The Adaptive Mind,” Project No. 533717223), the EU Horizon Europe Project “ARISE” under Grant 101135959 and the German Federal Ministry of Research, Technology and Space of Germany (BMFTR) Project “RIG” under Grant 16ME1001.}
\thanks{This work has been submitted to the IEEE for possible publication. Copyright may be transferred without notice, after which this version may no longer be accessible.}
}
\begin{document}

\maketitle
\thispagestyle{empty}
\pagestyle{empty}

\begin{abstract}

Bimanual manipulation requires reasoning about where to interact with an object and which arm should perform each action, a joint affordance localization and arm allocation problem that geometry-only planners cannot resolve without semantic understanding of task intent. Existing approaches either treat affordance prediction as coarse part segmentation or rely on geometric heuristics for arm assignment, failing to jointly reason about task-relevant contact regions and arm allocation. We reframe bimanual manipulation as a joint affordance localization and arm allocation problem and propose a hierarchical framework for task-aware bimanual affordance prediction that leverages a Vision-Language Model (VLM) to generalize across object categories and task descriptions without requiring category-specific training. Our approach fuses multi-view RGB-D observations into a consistent 3D scene representation and generates global 6-DoF grasp candidates, which are then spatially and semantically filtered by querying the VLM for task-relevant affordance regions on each object, as well as for arm allocation to the individual objects, thereby ensuring geometric validity while respecting task semantics. We evaluate our method on a dual-arm platform across nine real-world manipulation tasks spanning four categories: parallel manipulation, coordinated stabilization, tool use, and human handover. Our approach achieves consistently higher task success rates than geometric and semantic baselines for task-oriented grasping, demonstrating that explicit semantic reasoning over affordances and arm allocation helps enable reliable bimanual manipulation in unstructured environments.
\end{abstract}

\section{INTRODUCTION}
As robots are increasingly deployed in unstructured real-world environments, their ability to physically interact with and manipulate multiple objects in a task-aware manner becomes critical for effective task execution. A core aspect of such object interactions is 
the task-conditioned identification of object regions that support stable, functionally appropriate manipulation. 
In the context of bimanual manipulation, determining which arm should grasp which object region, namely a joint affordance localization and arm allocation problem, is therefore central to reliable bimanual execution to perform complex activities such as simultaneous object relocation, sorting, or collaborative manipulation. However, most existing approaches treat grasping as a single-arm, geometry-driven problem, ignoring the semantic reasoning needed to assign task-appropriate roles to each arm

\begin{figure}[th!]
    \centering
    \includegraphics[width=\linewidth]{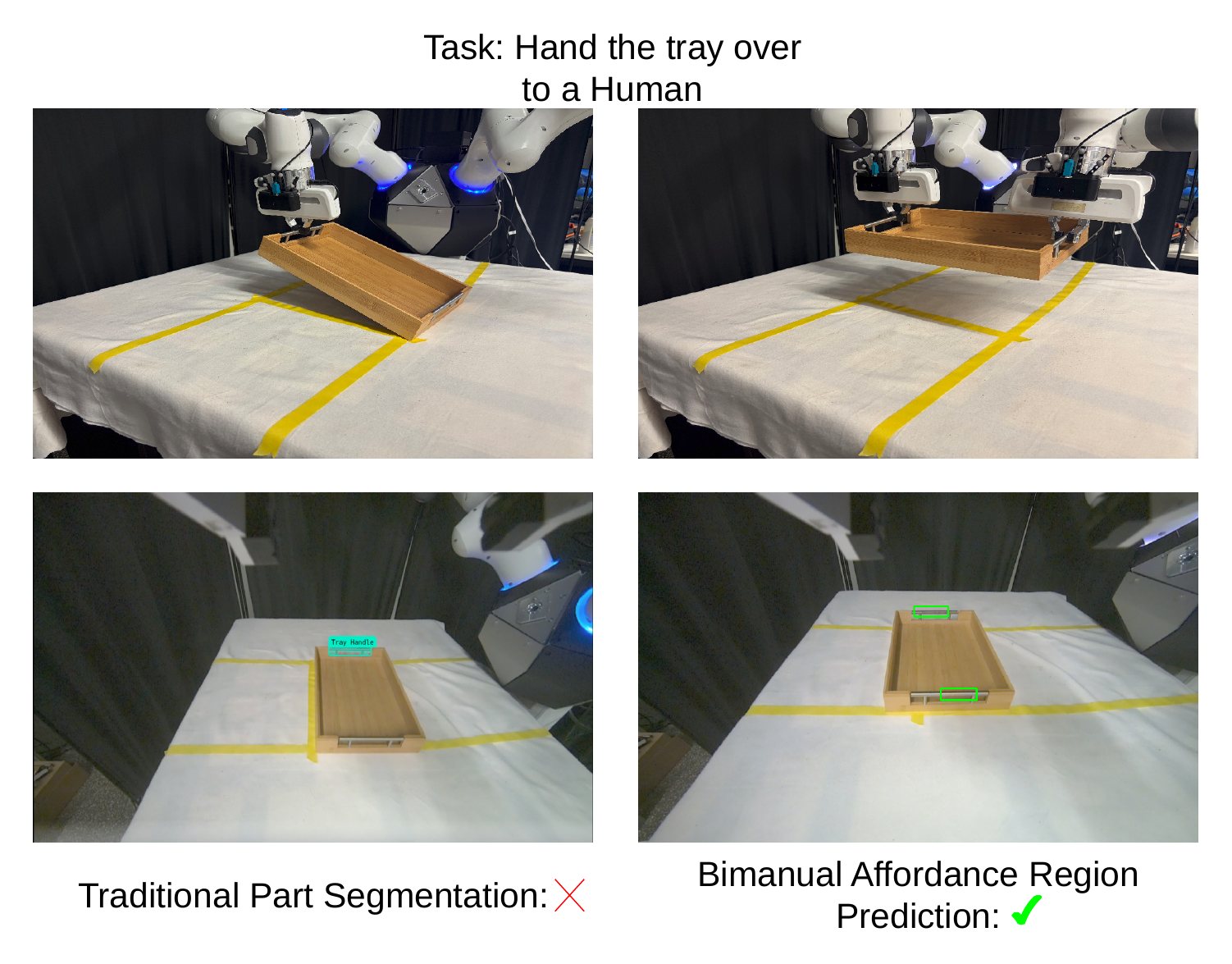}
    \caption{A motivating example: For the task “Hand the tray over to a human,” conventional labeling strategies typically reduce affordance annotation to coarse object-part segmentation for subsequent grasping. Such approaches fail to capture the task-specific requirement of simultaneously grasping both handles to ensure stability and safe transfer. In other words, traditional strategies overlook the coordinated, bimanual nature of the interaction and treat affordances as isolated object parts rather than functionally coupled regions.
    }
    \vspace{-1em}
    \label{fig:placeholder}
\end{figure}

Existing grasp generation methods~\cite{wang2022goal, urain2022se3dif,jauhri2024learning,karim2025dagdiff,karim2025dg16m} focus on geometry-driven candidate grasp synthesis for individual objects, while bimanual methods~\cite{liu2025leveraging,tang2023graspgpt,singh2024constrained} typically assume predefined object roles or task-specific training, or the availability of high-quality meshes, limiting their applicability in real-world multi-object contexts.
Recent vision-language models (VLMs) have shown strong semantic reasoning capabilities~\cite{Qwen3-VL,chen2024spatialvlm,kwon2024toward} and have been applied for open-vocabulary task-conditioned grasping in single-arm settings~\cite{tang2023graspgpt,mirjalili2023lan}. However, their use for coordinated bimanual affordance prediction, where semantic reasoning must jointly inform both affordance localization and arm allocation, remains unexplored.
In such contexts, achieving coordinated dual-hand operation presents several key challenges. First, robots must identify stable grasp configurations simultaneously for multiple objects, ensuring each arm achieves a stable, kinematically feasible grasp given workspace constraints and inter-arm proximity. Second, they must determine which object each hand should interact with, determining arm assignment based on task semantics, i.e. which object region is functionally appropriate for each arm given the intended action. Third, selecting appropriate grasp locations for each object is critical to optimize stability and efficacy while respecting reachability and avoiding inter-arm conflicts.

To address these limitations, we present a hierarchical bimanual manipulation framework in which a Vision-Language Model (VLM) provides task-aware affordance reasoning over novel objects and task descriptions, without category-specific training. Our approach queries a VLM to identify task-relevant contact regions on each object, grounding high-level task descriptions in 3D geometric grasp candidates to reason about coordinated, multi-object, dual-hand manipulation. By integrating high-level reasoning with geometry-aware grasp candidate generation, our method enables robots to plan stable, context-aware grasps for multiple objects simultaneously, expanding their capabilities in unstructured, open-world environments. 

In summary, our contributions are:
\begin{itemize}
    \item We reframe bimanual manipulation as a joint affordance localization and arm allocation problem, where task semantics drive how each arm interacts with the scene.
    \item A hierarchical framework integrating VLM-based affordance reasoning with geometric grasp generation for task-aware dual-arm execution, requiring no task-specific training.
    \item An arm allocation strategy derived from affordance regions that outperforms geometry-only and semantic baselines across nine real-world bimanual tasks.
\end{itemize}
We validate our approach on a dual-arm platform across nine real-world tasks and show that semantic affordance modelling yields significantly higher task success rates than geometry-only and semantic baselines.

\section{RELATED WORK}

The integration of large language models (LLMs) and vision-language models (VLMs) into robotic grasping has gained significant traction, driven by the promise of open-ended semantic understanding and zero-shot generalization~\cite{kim2024survey}. Lan-grasp~\cite{mirjalili2023lan} pioneered a zero-shot approach that chains an LLM, a VLM, and a grasp planner to achieve semantic grasp reasoning without task-specific training. Subsequent methods have extended this paradigm with large-scale datasets~\cite{vuong2024language}, visual grounding~\cite{lu2023vlgrasp}, fine-grained language instructions~\cite{sun2023flarg}, CLIP-based grasp synthesis in clutter~\cite{tziafas2023crog}, and open-vocabulary grasping of novel object categories~\cite{li2024ovgnet}. More advanced reasoning capabilities have also been explored, including implicit intent understanding~\cite{jin2025reasoning}, strategic decluttering via GPT-4o~\cite{qian2024thinkgrasp}, and generalizable task-oriented grasping with foundation models~\cite{tang2023graspclip,tang2025foundationgrasp}. On the geometric reasoning front, ShapeGrasp~\cite{li2024shapegrasp} decomposes objects into convex primitives for LLM-based part reasoning, while OWG~\cite{tziafas2024owg} combines VLMs with segmentation and grasp synthesis through zero-shot visual prompting. In the dexterous grasping domain, DexGYS~\cite{wei2024dexgys} and SayFuncGrasp~\cite{li2025language} extend language-guided grasping to multi-fingered hands, though both remain limited to single-hand manipulation. Liao et al.~\cite{liao2023decision} further demonstrate LLM-based decision-making for grasp selection in home-assistant scenarios.


A critical limitation shared by the majority of the aforementioned methods is that they mostly reduce grasp region identification to *object part selection*---the system identifies a named part (e.g., "handle") and generates grasps anywhere within that part's spatial extent~\cite{mirjalili2023lan, li2024shapegrasp, tang2025foundationgrasp, tang2023graspclip, tziafas2024owg, li2024ovgnet, lu2023vlgrasp, tziafas2023crog, sun2023flarg}. Liu et al.~\cite{liu2025leveraging} go a step further and achieve sub-part precision through visual prompts that divide the object surface into spatial grid regions for VLM reasoning. As noted by Heidinger et al.~\cite{heidinger2025twohanded}, this conflates the broader extent of a part with the localized region where functional contact actually occurs. Bimanual grasping introduces challenges beyond those of single-hand manipulation: determining the division of labor between hands, resolving spatial coupling constraints when two end-effectors interact with a shared object, and ensuring that affordance regions are jointly optimized, since the optimal region for one hand is conditioned on the other. This distinction is especially critical for bimanual grasping, where two grasp regions must be spatially compatible and jointly satisfy task constraints. 


The most closely related work is 2HandedAfforder~\cite{heidinger2025twohanded}, which trains a VLM-based affordance model on a large-scale dataset extracted from human activity videos via hand inpainting. While this represents an important advance in bimanual affordance understanding, it requires an extensive data generation and model training pipeline that incurs considerable computational and curation overhead. 
In contrast, our approach leverages the inherent reasoning capabilities of large VLMs in a zero-shot manner, circumventing task-specific dataset collection and training. Combined with region-based visual prompting, as done by~\cite{liu2025leveraging}, and the semantic reasoning as in~\cite{mirjalili2023lan}, our method extends these approaches to predict precise bimanual affordance regions and generate actionable task-oriented grasp configurations, offering a lightweight and readily deployable alternative for bimanual grasping in unstructured environments.


\section{PROPOSED APPROACH}\label{method}

\begin{figure*}[!h]
    \centering
    \includegraphics[width=\textwidth]{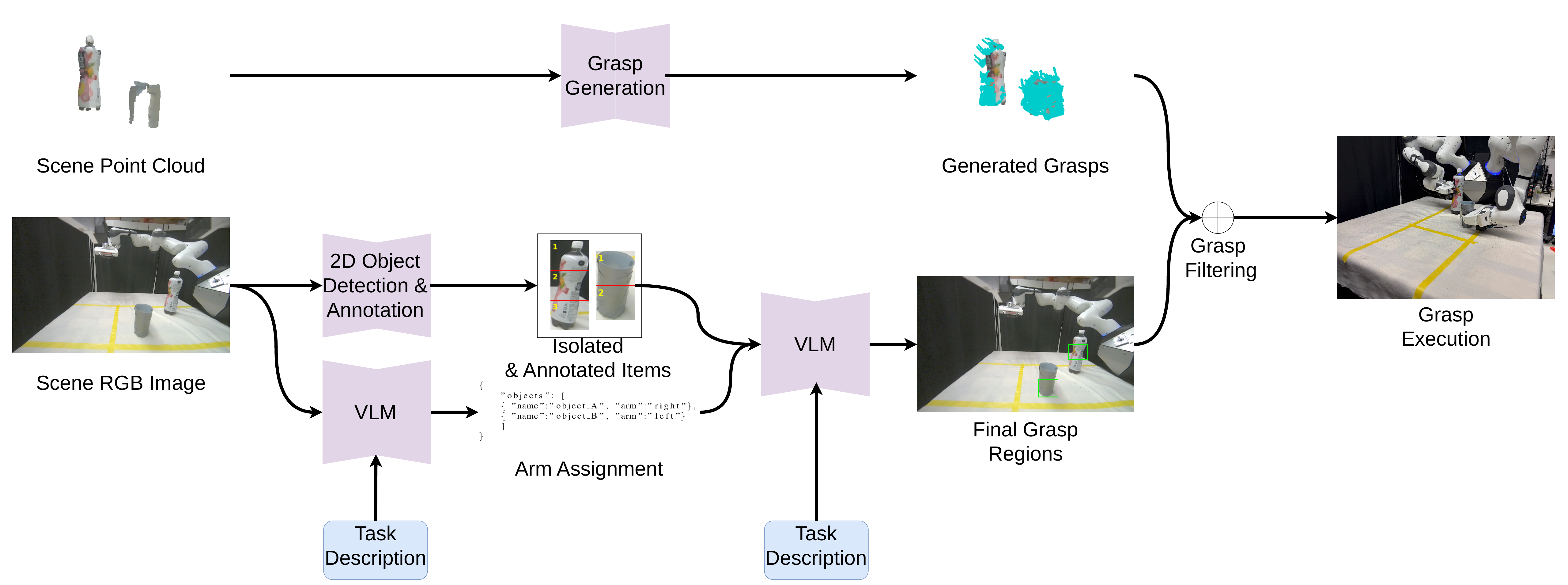}
    \caption{Overview of our proposed approach.
        Given a scene RGB image, a vision–language model (VLM) performs 2D object detection and annotation, followed by arm assignment and identification of final grasp regions based on the task description. In parallel, the RGB-D input is converted into a scene point cloud, from which candidate grasps are generated. These grasps are then filtered according to the grasp regions specified by the VLM, removing proposals that fall outside the designated areas. The selected grasps which align with the task-relevant parts of each object are then executed.
        }
    \label{fig:overview}
    \vspace{-1em}
\end{figure*}

Our grasp planning architecture integrates high-level semantic reasoning with robust geometric sampling to enable sophisticated bimanual manipulation. The pipeline is divided into three primary stages: multi-view scene representation, hierarchical semantic reasoning, and candidate filtering for bimanual execution.

\subsection{Multi-View Scene Representation and Global Sampling}\label{representation}

To ensure a comprehensive representation of the workspace and mitigate occlusions, we employ a multi-view sensing strategy. Two RGB-D sensors are strategically positioned to provide overlapping fields of view. The depth data and color information from both sensors are captured synchronously, transformed into a unified robot base coordinate frame, and fused into a single, dense 3D point cloud. By merging these multiple perspectives, we produce a more complete representation of the environment than is possible with a single-view system.

The resulting merged point cloud serves as the input for a grasp generation algorithm. Importantly, any off-the-shelf grasp generator can operate directly on this data without requiring additional training, making the pipeline immediately usable with existing tools. This model evaluates the global geometry of the scene to identify a wide array of 6-DOF candidate grasp poses across the entire workspace, considering surface normals and collision geometries. By sampling the entire workspace globally at the start of the pipeline, the algorithm generates a robust set of candidates that account for the full environmental context and potential inter-object collisions.

\subsection{Hierarchical Semantic Reasoning}

We then employ a hierarchical reasoning stage using a Vision-Language Model (VLM) to refine the global grasp pool. This process proceeds in three main steps: arm allocation, object-specific region selection, and post-processing for execution.

\noindent\textbf{Step 1: Arm Allocation and Task Strategy}  

The VLM is provided with a task description and a list of target objects. It determines for each object whether a single-arm grasp (\texttt{left} or \texttt{right}) is sufficient or if a coordinated \texttt{bimanual} grasp (both arms for a single object) is required based on the object's perceived scale and the task's stability requirements.

The prompt given to the VLM for arm allocation is:  

\begin{quote}
\ttfamily
You are an intelligent bimanual robot. Your task is to assign "left", "right", or "bimanual" strategies to the following objects: \{objects\}.\\
Consider the requirements for the task: \{task\_description\}.\\
Use "bimanual" if a single object is heavy or wide; otherwise, assign "left" or "right" to ensure coordinated dual-object manipulation.\\
\\
Your output must be a single JSON object: \{"objects": [\{"name": "<obj>", "arm": "left/right/bimanual"\}]\}. 
\end{quote}  

\begin{figure}
    \centering
    \includegraphics[width=\linewidth]{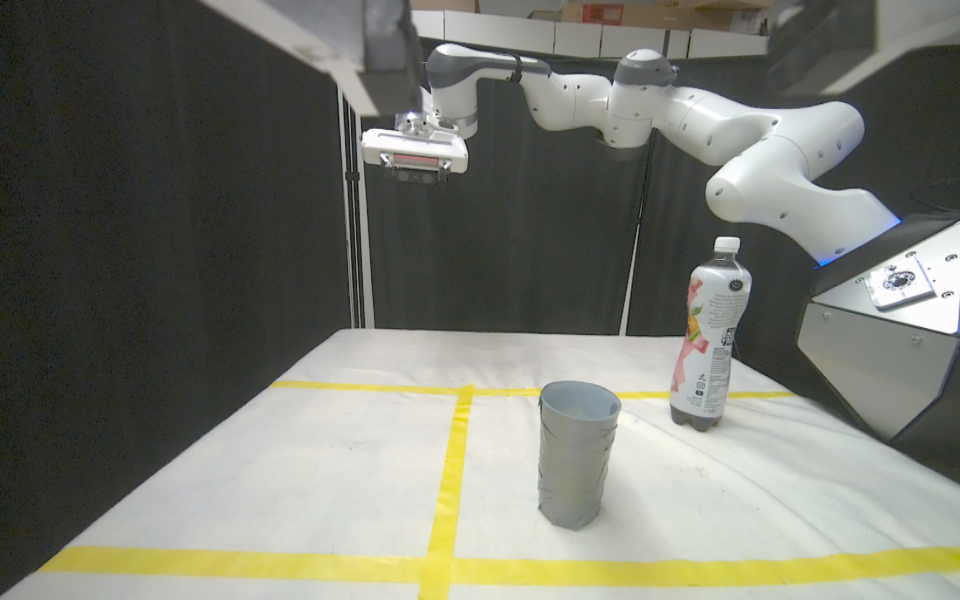}
    \caption{The figure illustrates an example RGB input to our pipeline. In this scenario, the robot is tasked with grasping two objects: a bottle and a cup.}
    \label{fig:example-scene}
    \vspace{-1em}
\end{figure}

An example output for the scene depicted in Figure \ref{fig:example-scene} from the VLM for Step 1 is shown in Listing~\ref{lst:arm-json}:  

\begin{lstlisting}[caption={Example JSON output for arm allocation},label={lst:arm-json}]
{
    "objects": [        
    { "name":"bottle", "arm":"right"},
    { "name":"cup", "arm":"left"}
    ]
}
\end{lstlisting}  

\noindent\textbf{Step 2: Object-Specific Grasp Planning}  

Once the grasp allocation is determined, we isolate target objects using a 2D detection pipeline. Following a size-adaptive approach, a grid is overlaid on the detected region. A similar idea has been explored by the work of Huang et al.~\cite{huang2024combining}, though their approach is limited to single-object scenes and does not dynamically adjust the grid size. The number of grid cells is dynamically computed from the object’s physical dimensions estimated from the multi-view depth data, thereby ensuring that each cell approximately matches the scale of the robot’s gripper. The VLM is then prompted to select specific grid cells representing optimal grasping regions.

The prompt for object-specific grasp planning is:  

\begin{quote}
\ttfamily
The image shows a \{object1\}. You are a bimanual robot performing: \{task\_description\}.\\
You are assigned the \{grasp\_arm\_object1\} strategy for this \{object1\}.\\
If the strategy is "bimanual", you MUST select two distinct grid cells (one for each hand) on opposite sides of the object for balance. If "left" or "right", select one cell for that arm.\\
\\
JSON Output Format: \\
\{\\
\ \ "objects": ["\{object1\}"],\\
\ \ "num\_hands\_robot": \{hands\_count\},\\
\ \ "cell\_robot\_right": ["<cell\_number>"],\\
\ \ "cell\_robot\_left": ["<cell\_number>"]\\
\}
\end{quote}  

\begin{figure}
    \centering
    \vspace{-1em}
    \includegraphics[scale=1]{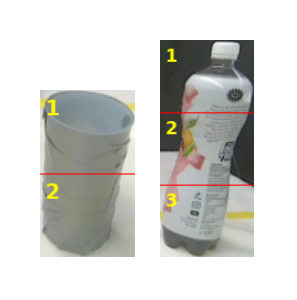}
    \vspace{-1em}
    \caption{This figure presents an example of the cropped and annotated images provided to the VLM for grasp region assignment. Here, the inputs correspond to a cup and a bottle.}
    \label{fig:example-annotation}
    \vspace{-1em}
\end{figure}

An example output for one the items depicted in Figure \ref{fig:example-annotation} for a coordinated bimanual strategy is shown in Listing~\ref{lst:grasp-json}:  

\begin{lstlisting}[caption={Example JSON output for coordinated grasp regions},label={lst:grasp-json}]
{
    "objects": ["bottle"],
    "num_hands_robot": 1,
    "cell_robot_right": ["2"],
    "cell_robot_left": []
}
{
    "objects": ["cup"],
    "num_hands_robot": 1,
    "cell_robot_right": [],
    "cell_robot_left": ["2"]
}
\end{lstlisting} 

\subsection{Post-Processing and Bimanual Execution}

In the final stage, the 2D grid cells selected by the VLM are projected back into 3D space using camera intrinsics and the current transformation between the sensors and the robot's base. This isolates the 3D points corresponding to the VLM's selected regions, which are then used to filter the global grasp pool. Any candidate pose whose center points fall outside these selected 3D regions is discarded.

The pipeline branches based on the VLM's strategy:
\begin{itemize}
    \item \textbf{Parallel Mode:} If different objects are assigned to different arms, the system independently selects the highest-scoring collision-free candidate—where the score is computed by the grasp generation algorithm—within each object’s designated region.
    \item \textbf{Coordinated Mode:} If two arms are assigned to a single object, the system performs a pairwise search between the filtered left and right grasp pools to find a pair $(g_L, g_R)$ that maximizes a combined quality score while maintaining a minimum safety distance ($||pos(g_L) - pos(g_R)|| > d_{min}$) to prevent gripper-to-gripper collision.
\end{itemize}

The final targets are refined through approach-axis alignment and pre-grasp computation before being dispatched to a bimanual motion controller. This hierarchical strategy ensures that the resulting robot motion is both semantically aligned with the high-level task and geometrically feasible within the 3D environment.

\section{EXPERIMENTAL EVALUATION}
The goal of our experimental evaluation is to determine whether explicitly modeling language-conditioned bimanual affordances leads to manipulation behavior that is not only geometrically feasible, but also semantically correct and strategy-consistent. While many grasping systems are evaluated based on physical stability or grasp success rate, such metrics do not capture whether a robot used the right arm, the right number of arms, or grasped the right regions for a given task. Our experiments therefore focus on measuring alignment between executed grasps and human-defined manipulation strategies.

\begin{figure*}[!t]
    \centering
    \subfloat[Stool handover: The robot performs a symmetric bimanual grasp to lift and present the stool while preserving accessible regions for the human collaborator.]{%
        \includegraphics[width=0.24\textwidth]{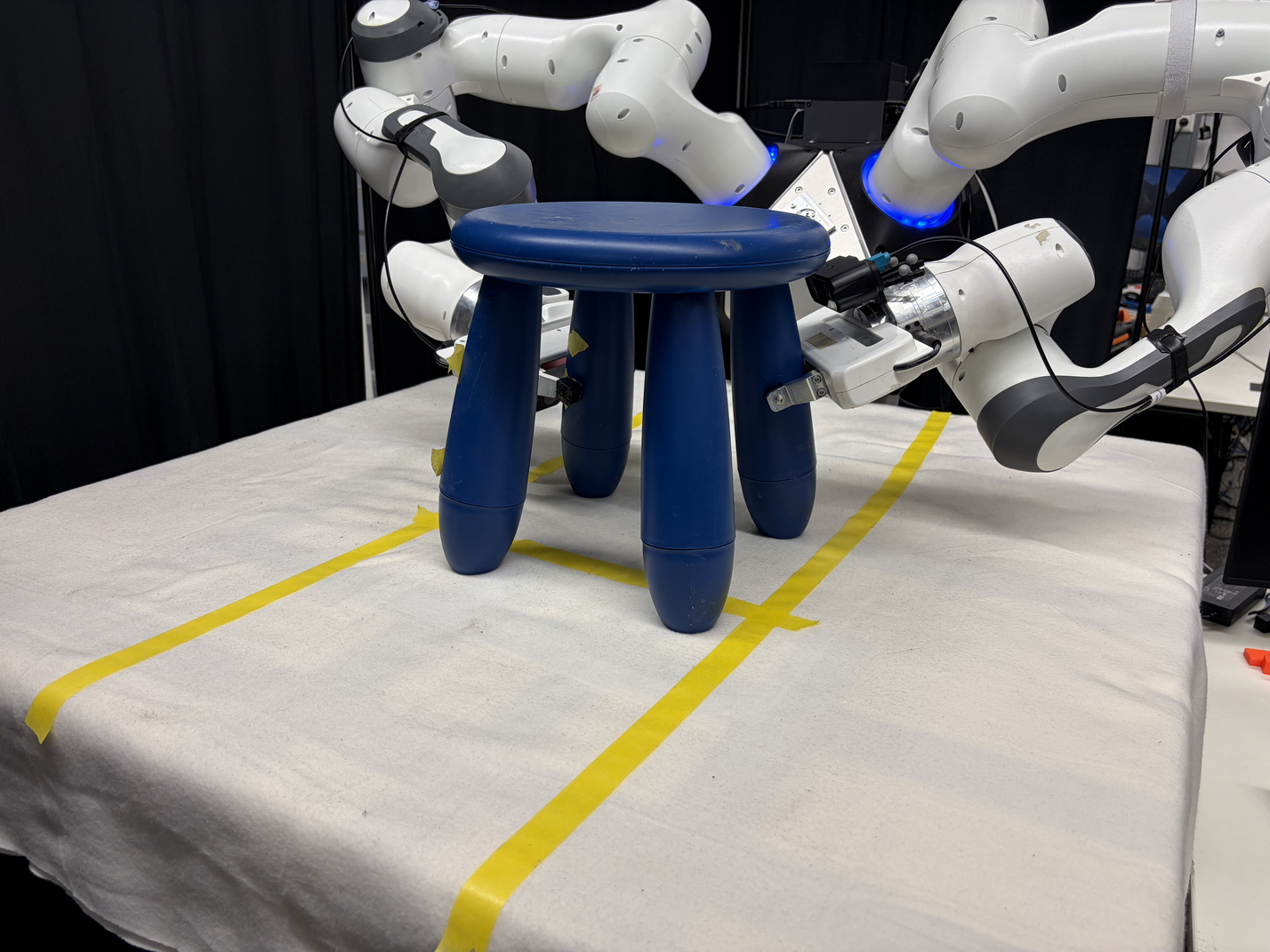}%
        \label{fig:subfig1}%
    }
    \hfill
    \subfloat[Bottle opening: One arm stabilizes the bottle body while the other grasps and unscrews the cap, demonstrating complementary role assignment and part-level affordance selection.]{%
        \includegraphics[width=0.24\textwidth]{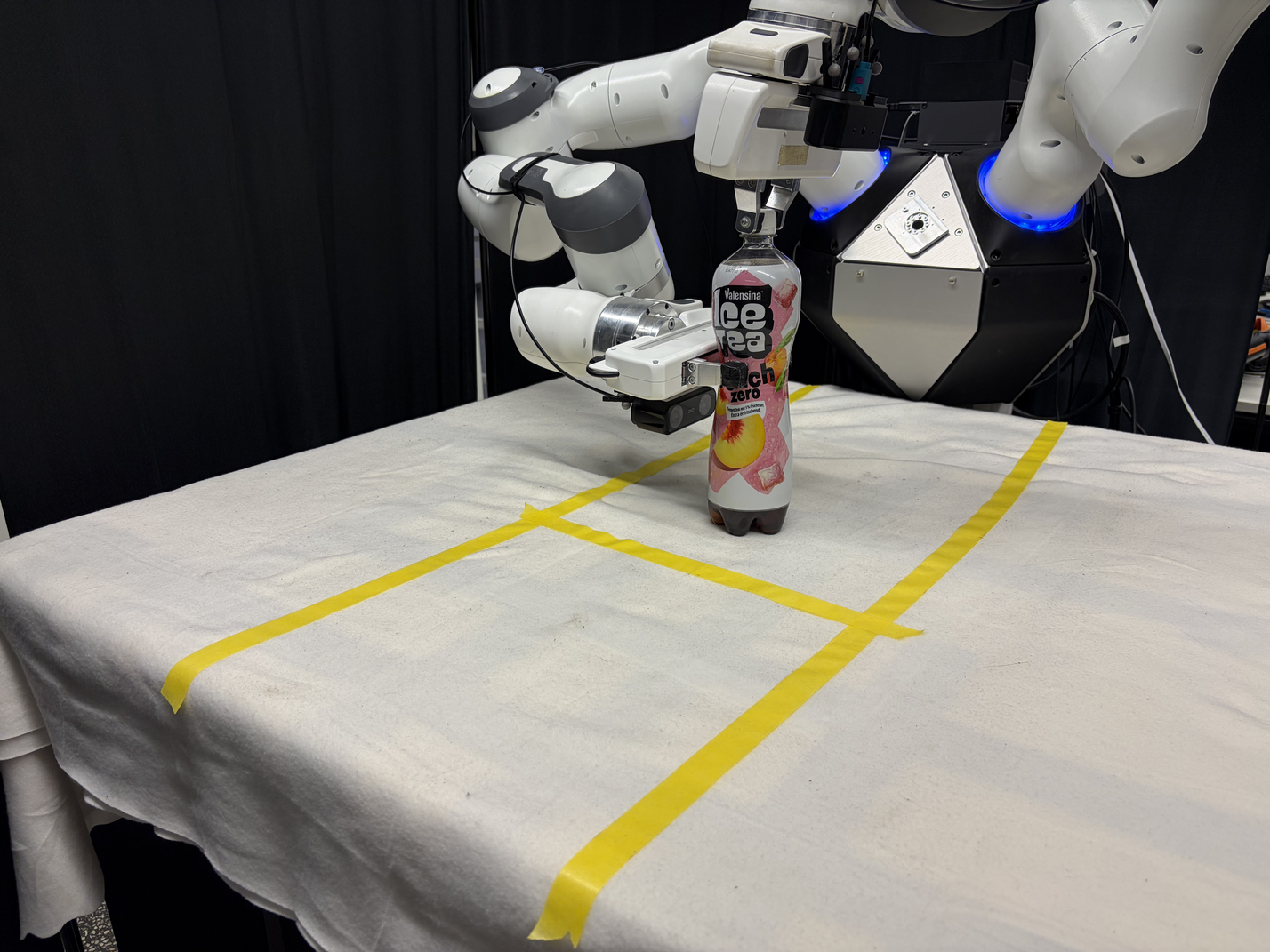}%
        \label{fig:subfig2}%
    }
    \hfill
    \subfloat[Tray handover: Coordinated dual-arm grasping of the tray ensures balanced support and stable transfer during human–robot interaction.]{%
        \includegraphics[width=0.24\textwidth]{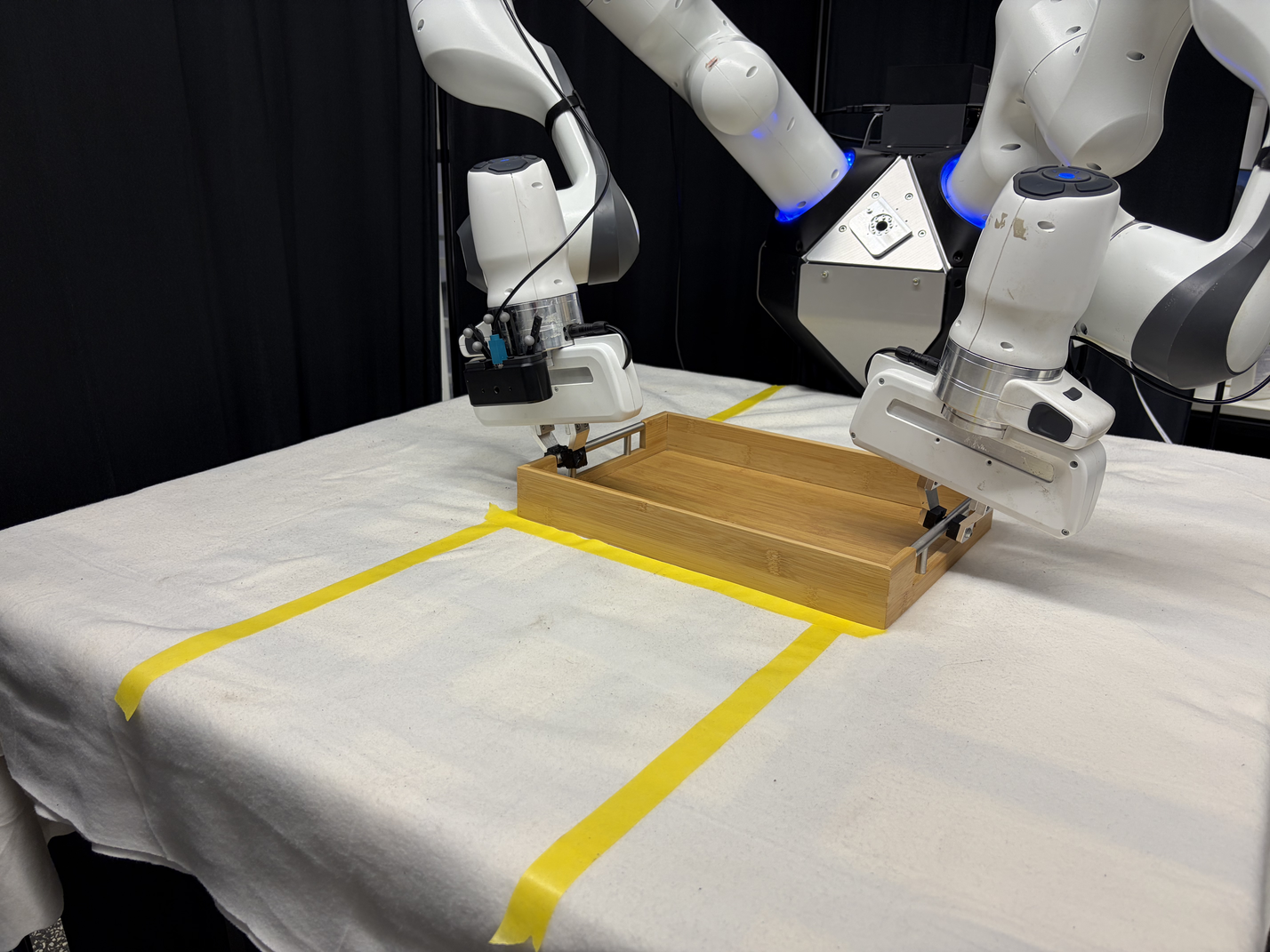}%
        \label{fig:subfig3}%
    }
    \hfill
    \subfloat[Cooking pot opening: The robot assigns complementary roles, with one arm stabilizing the pot body and the other removing the lid.]{%
        \includegraphics[width=0.24\textwidth]{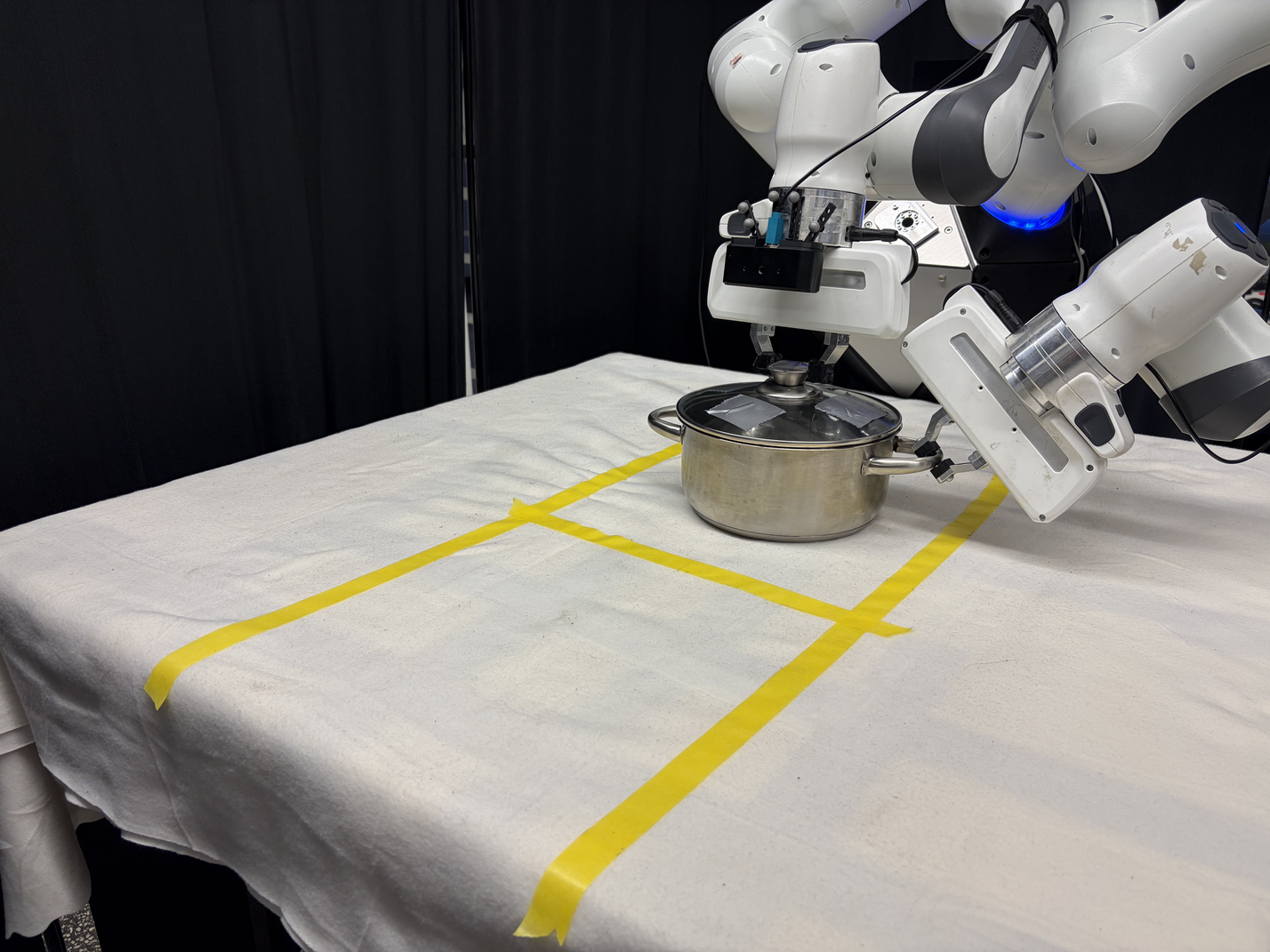}
        \label{fig:subfig4}%
    }
    \caption{Representative successful executions of our language-conditioned hierarchical grasp planning framework. The examples illustrate symmetric large-object stabilization (\ref{fig:subfig1}, \ref{fig:subfig3}), complementary bimanual manipulation (\ref{fig:subfig4}), and asymmetric role assignment with fine-grained affordance selection (\ref{fig:subfig2}).}
    \label{fig:results}
\end{figure*}

\subsection{Experimental Setup}

We evaluate our language-conditioned bimanual grasp planning framework on a dual-arm robotic platform composed of two Franka Emika Panda manipulators\footnote{\url{https://www.franka.de/panda}} arranged to provide overlapping workspaces over a shared tabletop environment. This configuration enables both independent single-arm manipulation and coordinated dual-arm interaction within a common scene.

\textbf{Multi-View Perception.}
Following the multi-view representation strategy described in Section \ref{representation}, we deploy two ZED X Mini stereo RGB-D cameras\footnote{\url{https://www.stereolabs.com/en-us/store/products/zed-x-mini-stereo-camera}} positioned to provide complementary and overlapping views of the workspace. The synchronized RGB and depth streams are transformed into the robot base frame via calibrated extrinsics and fused into a unified dense 3D point cloud.

The multi-view setup reduces perception failures caused by self-occlusions or large objects and yields a more complete geometric representation of the environment. This is particularly important in cluttered, multi-object scenes where incomplete geometry can negatively affect both grasp candidate generation and downstream semantic filtering.

\textbf{Global Grasp Candidate Generation.}
We generate a dense set of 6-DoF grasp hypotheses from the fused point cloud using AnyGrasp~\cite{fang2023anygrasp}, a learning-based grasp detection method capable of category-agnostic grasp prediction in cluttered scenes. AnyGrasp operates in a task-agnostic manner and proposes collision-aware grasp candidates across all visible objects. This stage ensures that downstream semantic reasoning operates over a geometrically feasible and diverse grasp pool.

\textbf{Language-Conditioned Hierarchical Filtering.}
The global grasp set is refined through the hierarchical reasoning module introduced in Section \ref{method}. First, GPT-5 performs arm allocation, determining whether each object should be manipulated with the left arm, right arm, or both arms based on task semantics and perceived object characteristics. Second, the GPT-5 selects object-specific grasp regions using grid-based prompting. The selected regions are projected into 3D space and used to filter the global grasp candidates.

\subsection{Experimental Tasks}

We evaluate the proposed method on a diverse set of real-world manipulation tasks designed to investigate distinct components of the system, including arm allocation reasoning, object-to-hand assignment, part-level affordance selection, coordinated dual-arm stabilization, and human-aware grasping.

\begin{itemize}

    \item \textbf{Stir vegetables.}  
    One arm stabilizes a container while the other manipulates a fork. This task evaluates parallel multi-object manipulation and role assignment for a dynamic interaction.

    \item \textbf{Pick up pot.}  
    The robot grasps and lifts a pot from the table. This task evaluates arm allocation decisions conditioned on object size and stability requirements.

    \item \textbf{Open pot.}  
    One arm holds the pot body while the other removes the lid. This task evaluates coordinated manipulation of two related objects with complementary roles.

    \item \textbf{Open bottle.}  
    One arm stabilizes the bottle while the other unscrews the cap. This task evaluates fine-grained part-level region selection and asymmetric arm assignment.

    \item \textbf{Pour from bottle to cup.}  
    The robot grasps a bottle and performs a controlled pouring motion.  
    This task evaluates task-aware grasp region selection that preserves functional affordances during object reorientation.

    \item \textbf{Sweeping with hand broom.}  
    The robot grasps a broom and executes planar sweeping motions.  
    This task evaluates zero-shot affordance reasoning for tool usage.

    \item \textbf{Drill machine grasp.}  
    The robot detects and grasps a drill machine.  
    This task evaluates handle-oriented grasp selection for tool-like objects without task-specific training.

    \item \textbf{Tray handover (robot-to-human).}  
    The robot grasps a tray and presents it to a human collaborator. This task evaluates coordinated bimanual stabilization while maintaining accessible handover regions.

    \item \textbf{Stool handover (robot-to-human).}  
    The robot lifts and presents a stool for handover. This task evaluates large-object bimanual grasp pairing, symmetric region selection, and safe human interaction.

\end{itemize}

Collectively, these tasks span unimanual grasping, parallel multi-object manipulation, coordinated dual-arm stabilization, tool-oriented affordance reasoning, and human-interactive handover. This diversity enables evaluation of both geometric robustness and semantic generalization under zero-shot language conditioning.

\subsection{Baselines and Ablation Studies}

To quantify the contribution of individual components, we compare our full framework against structured baselines and ablations. All methods share identical hardware, perception, multi-view fusion, and AnyGrasp-based global grasp generation. Variants differ only in their semantic reasoning and filtering mechanisms.

\subsubsection{ Geometry-Only Baseline}

This baseline removes all language-conditioned reasoning.

\begin{itemize}
    \item No arm allocation via VLM
    \item No region-level filtering
    \item Grasp selection purely based on AnyGrasp scores
\end{itemize}

Arm assignment is determined heuristically (nearest arm), and a size-based threshold is used to trigger bimanual grasps for large objects. The bimanual case is treated as the coordinated mode as described in Section \ref{method}

\textbf{Objective.}  
Isolates the effect of semantic reasoning by evaluating purely geometry-driven grasp selection.

\subsubsection{ Arm Allocation Only (No Region Filtering)}

This ablation retains language-conditioned arm allocation but removes part-level region selection.

\begin{itemize}
    \item VLM-based left/right/bimanual strategy retained
    \item No grid-based region filtering
    \item Highest-scoring grasp selected within assigned object
\end{itemize}

\textbf{Objective.}  
Evaluates whether high-level arm allocation alone suffices without explicit affordance-region reasoning.

\subsubsection{ Region Selection Only (No Arm Allocation)}

This variant preserves grid-based region selection while replacing arm allocation with heuristic assignment.

\begin{itemize}
    \item VLM-based region selection retained
    \item Arm assignment determined by proximity
    \item No language-conditioned coordination strategy
\end{itemize}

\textbf{Objective.}  
Assesses the necessity of coordinated arm reasoning beyond part-level grasp selection.

\subsubsection{ VLPart Baseline}

We additionally compare against a semantic part-segmentation baseline using VLPart~\cite{sun2023going}.

\begin{itemize}
    \item Open-vocabulary part segmentation via VLPart
    \item Grasp candidates filtered using detected part regions
    \item Arm assignment determined heuristically
\end{itemize}

VLPart provides semantic part localization but does not perform task-conditioned arm allocation or hierarchical reasoning over manipulation strategy.

\textbf{Objective.}  
Evaluates whether generic part segmentation is sufficient for functional grasping, compared to our fully task-conditioned hierarchical framework.

\subsection{Results}

We evaluate all methods with respect to \emph{grasp strategy alignment} rather than purely geometric grasp stability. Since the core contribution of our framework lies in language-conditioned reasoning over arm allocation and grasp region selection, performance is measured by how closely each method’s execution matches a human-defined ground-truth grasping strategy.

For each task, a human annotator specifies the expected manipulation strategy, including:
\begin{itemize}
    \item The required arm configuration (left, right, or bimanual),
    \item The number of hands involved,
    \item The appropriate grasp region(s) on the object.
\end{itemize}

A trial is considered successful if the executed grasp satisfies \emph{all} strategy criteria. For bimanual tasks, both arm assignment and region symmetry constraints must be met. For tool-based tasks, grasps must occur on semantically appropriate functional regions (e.g., handle versus non-functional surface). For human handover tasks, the selected regions must preserve accessibility for the human collaborator. This evaluation protocol directly measures alignment with human-intended manipulation strategy rather than low-level grasp feasibility.

Table~\ref{tab:strategy_results} reports the strategy alignment rate (\%) for all methods across nine manipulation tasks. For each task, each method is evaluated with 10 different object positions. Four representative successful grasp executions are shown in Figure \ref{fig:results}.
\begin{table*}[h]
\centering
\caption{Strategy Alignment Rate (\%) across tasks. 
Tasks are grouped by manipulation type.}
\label{tab:strategy_results}
\begin{tabular}{lccccccccc|c}
\toprule
\textbf{Method} 
& \multicolumn{3}{c}{\textbf{Parallel / Coordination}} 
& \multicolumn{4}{c}{\textbf{Affordance-Sensitive}} 
& \multicolumn{2}{c}{\textbf{Large Object}}  
& \textbf{Mean} \\
\cmidrule(lr){2-4} \cmidrule(lr){5-8} \cmidrule(lr){9-10}
& Stir 
& Pick Pot 
& Open Pot 
& Open Bottle 
& Pour 
& Sweep 
& Drill
& Tray  
& Stool 
&  \\
\midrule
Geometry-Only 
& 10 & 20 & 0 
& 0 & 0 & 10 
& 0 & 10 
& 30 
& 9.0 \\

Arm Only 
& 20 & 80 & 20 
& 20 & 10 & 80 
& 70 & 30 
& 80 
& 45.6 \\

Region Only 
& 80 & 20 & 80 
& 80 & 80 & 20 
& 30 & 80 
& 30 
& 55.6 \\

VLPart 
& 60 & 50 & 60 
& 50 & 60 & 50 
& 50 & 60 
& 60 
& 55.6 \\

\textbf{Ours} 
& \textbf{100} & \textbf{80} & \textbf{90} 
& \textbf{90} & \textbf{80} & \textbf{100} 
& \textbf{70} & \textbf{90} 
& \textbf{100} 
& \textbf{88.9} \\

\bottomrule
\end{tabular}
\vspace{-1em}
\end{table*}
\subsection{Overall Performance}

Our method achieves the highest mean strategy alignment rate (88.9\%), substantially outperforming Geometry-Only (9.0\%), Arm Only (45.6\%), Region Only (55.6\%), and VLPart (55.6\%). The Geometry-Only baseline collapses across nearly all tasks, indicating that geometric feasibility alone is insufficient for strategy-consistent manipulation.

In contrast, the proposed hierarchical framework maintains consistently high performance across all task categories. This demonstrates that combining language-conditioned arm allocation with region-level affordance reasoning is necessary for reliable execution of semantically correct bimanual grasping strategies.

\subsection{Coordination-Dominant Tasks}

For parallel and coordination-heavy tasks (Stir, Pick Pot, Open Pot), different ablations exhibit complementary strengths and weaknesses. Arm Only performs strongly when global arm assignment dominates (Pick Pot: 80\%) but fails in tasks requiring correct region-level role distribution (Stir: 20\%, Open Pot: 20\%). Conversely, Region Only performs well when correct contact regions are critical (Stir: 80\%, Open Pot: 80\%) but struggles when appropriate arm assignment is decisive (Pick Pot: 20\%). In the case of Pick Pot, this failure is largely attributable to inaccurate size estimation from the fused point cloud: if the reconstructed geometry underestimates the pot’s true extent due to occlusion or depth noise, the heuristic arm assignment may incorrectly select a single-arm strategy, despite the task requiring stable bimanual lifting. VLPart remains moderate but does not consistently capture coordination structure (50--60\%). Our method achieves 100\%, 80\%, and 90\% respectively, indicating that both coordinated arm reasoning and region-level affordance selection are required for consistent strategy alignment..

\subsection{Affordance-Sensitive Tasks}

In affordance-sensitive and tool-oriented scenarios (Open Bottle, Pour, Sweep, Drill), Region Only generally outperforms Arm Only in tasks dominated by functional contact selection (Open Bottle: 80\% vs.\ 20\%, Pour: 80\% vs.\ 10\%). In these cases, identifying the correct functional region (e.g., bottle cap, pouring body) is the primary determinant of success, and once the appropriate contact area is selected, the specific arm choice plays a comparatively minor role.

However, this trend reverses in Sweep (20\% vs.\ 80\%) and Drill (30\% vs.\ 70\%), where Arm Only significantly outperforms Region Only. A plausible explanation is that, for elongated or handle-based tools, the geometrically most stable grasp locations often coincide with the semantically correct ones (i.e., the handle). As a result, even without explicit region filtering, the global grasp generator frequently proposes stable handle grasps, reducing the marginal benefit of region-level reasoning. In contrast, correct arm assignment becomes more influential: sweeping and drilling implicitly impose directional and workspace constraints, such as dominant-side motion, collision avoidance with the environment, or sufficient range of motion for forward tool actuation. A heuristic arm choice based solely on proximity can therefore select an arm with limited workspace or unfavorable kinematic configuration, leading to strategy misalignment despite plausible contact regions. Explicit arm reasoning helps ensure that the selected manipulator can execute the intended motion effectively.

VLPart achieves relatively stable but moderate performance (50--60\%) across these tasks, suggesting that generic part segmentation provides useful structural cues but lacks task-conditioned coordination reasoning. Our method achieves 70--100\% across all affordance-sensitive tasks, demonstrating that integrating region-level affordance modeling with explicit arm allocation yields consistently superior strategy alignment.

\subsection{Human-Interactive and Large-Object Tasks}

In robot-to-human handover scenarios involving large-object stabilization (Tray and Stool), coordinated dual-arm reasoning becomes critical. \textit{Arm Selection Only} performs strongly in symmetric stabilization cases (Stool: 80\%) but lacks consistent region-level reasoning. \textit{Region Selection Only} underperforms due to heuristic arm assignment (Tray: 30\%, Stool: 30\%). VLPart again exhibits moderate performance (50--60\%) but does not enforce symmetric or socially appropriate dual-arm placement. Our method achieves 70\% (Tray) and 100\% (Stool), indicating improved alignment with human-intended bimanual affordances, particularly in large-scale and human-interactive manipulation scenarios.

\section{DISCUSSION}

The results demonstrate that evaluating manipulation through \emph{grasp strategy alignment} reveals structural differences between geometric grasping, part segmentation, and coordinated affordance reasoning. Because success is defined by agreement with human-specified arm configuration, number of hands, and grasp regions, our evaluation isolates whether a method captures the intended manipulation strategy rather than merely achieving physical stability. Under this criterion, geometry-only grasp selection collapses, confirming that collision-free 6-DoF grasps alone are insufficient for semantically correct execution.

The ablation results expose a structured decomposition of the bimanual manipulation problem. Arm allocation alone performs strongly in tasks where global coordination decisions dominate (e.g., Pick Pot, Sweep, Drill), but fails in scenarios where fine-grained functional contact selection is critical (e.g., Open Bottle, Pour). Conversely, region-only reasoning succeeds in tasks where identifying the correct functional surface determines success (e.g., Stir, Open Pot, Open Bottle, Pour), yet fails when correct arm configuration depends on accurate size estimation or symmetric stabilization requirements (e.g., Pick Pot, Stool). These complementary behaviors indicate that neither global arm assignment nor local region prediction alone is sufficient across task categories.

The comparison with VLPart further clarifies this distinction. While semantic part segmentation provides moderate performance across tasks, it remains object-centric and does not explicitly model interaction structure. In large-object handover tasks such as Tray and Stool, successful execution depends not merely on selecting plausible parts, but on enforcing symmetric, accessible, and task-consistent contact regions relative to both arms. As reflected in the results, segmentation-based filtering often selects reasonable object parts but fails to enforce correct arm configuration or coordination symmetry, leading to strategy mismatches.

By contrast, our hierarchical framework explicitly models bimanual affordances at both levels: determining how many arms to use and which arms to assign, and selecting task-consistent grasp regions conditioned on that allocation. The consistently higher strategy alignment across coordination-heavy, affordance-sensitive, tool-oriented, and human-interactive tasks suggests that this interaction-centric formulation generalizes more robustly than geometry-only or part-based approaches.

Overall, these findings support a shift in perspective for dual-arm manipulation. Effective bimanual systems must reason over coordinated affordances that integrate task intent, arm configuration, physical object scale, and spatial symmetry. The proposed language-conditioned hierarchical approach provides a structured mechanism to achieve this integration, enabling zero-shot strategy alignment in open-world manipulation scenarios.

\section{CONCLUSION}

We presented a hierarchical framework for task-aware bimanual affordance prediction that reframes dual-arm manipulation as a joint affordance localization and arm allocation problem. By integrating VLM-based semantic reasoning with geometric grasp candidate generation, our approach determines task-relevant contact regions and arm assignments in a zero-shot manner, without requiring task-specific training or fine-tuning. Experiments across nine real-world manipulation tasks demonstrate that our method 
substantially outperforms geometry-only and purely semantic baselines.
The ablation results reveal a clear complementarity between arm allocation and region selection: neither component alone is sufficient across the full range of task categories, confirming that both levels of reasoning are necessary for reliable bimanual execution.

Several limitations remain. The pipeline relies on accurate object detection and performance degrades under heavy occlusion. Arm allocation is currently determined by the VLM based on perceived object scale and task description, without explicit kinematic feasibility checking at the allocation stage. Additionally, execution currently operates in a quasi-static regime and does not account for dynamic scenes or reactive replanning. Future work will explore closing these gaps through kinematic-aware arm allocation, tighter integration of grasp feasibility feedback into the VLM reasoning loop, and extension to mobile manipulation platforms where workspace constraints are less fixed. 






\bibliographystyle{IEEEtran}
\bibliography{references}

\end{document}